\documentclass{article}

    \PassOptionsToPackage{numbers, compress}{natbib}
    \usepackage[preprint]{neurips_2019}

\usepackage[utf8]{inputenc} % allow utf-8 input
\usepackage[T1]{fontenc}    % use 8-bit T1 fonts
\usepackage{hyperref}       % hyperlinks
\usepackage{url}            % simple URL typesetting
\usepackage{booktabs}       % professional-quality tables
\usepackage{amsfonts}       % blackboard math symbols
\usepackage{nicefrac}       % compact symbols for 1/2, etc.
\usepackage{microtype}      % microtypography

\usepackage{microtype}
\usepackage{graphicx}
\usepackage{booktabs} % for professional tables
\usepackage{amsmath}
\usepackage[dvipsnames]{xcolor}

\usepackage{subcaption}
\usepackage{wrapfig}

\title{QPyTorch: A Low-Precision Arithmetic\\ Simulation Framework}

\author{%
  Tianyi Zhang, Zhiqiu Lin, Guandao Yang, Christopher De Sa \\
  Cornell University, Department of Computer Science\\
  \texttt{\{tz58,zl279,gy46,cmd353\}@cornell.edu}\\
}

\begin{document}

\maketitle

\begin{abstract}
Low-precision training reduces computational cost and produces efficient models. 
Recent research in developing new low-precision training algorithms often relies on simulation to empirically evaluate the statistical effects of quantization while avoiding the substantial overhead of building specific hardware.
To support this empirical research, we introduce QPyTorch, a low-precision arithmetic simulation framework.
Built natively in PyTorch, QPyTorch provides a convenient interface that minimizes the efforts needed to reliably convert existing codes to study low-precision training.
QPyTorch is general, and supports a variety of combinations of precisions, number formats, and rounding options. 
Additionally, it leverages an efficient fused-kernel approach to reduce simulator overhead, which enables simulation of large-scale, realistic problems.
QPyTorch is publicly available at \url{https://github.com/Tiiiger/QPyTorch}.
\end{abstract}

\section{Introduction}
Low-precision arithmetic is one of the most successful techniques in compressing and accelerating Deep Neural Networks (DNNs). 
To obtain high-performance low-precision models, many works study low-precision training algorithms~\cite{binary-net, binaryconnect, WAGE}. 
Along with the advent of low-precision training techniques, there has been a growing interest in software that simulates low-precision computation in order to understand its statistical effects on training.
This simulation circumvents the overhead of building hardware while enabling investigations into the numerical behaviors of different designs.

To facilitate empirical research on low-precision training, we introduce QPyTorch, a low-precision arithmetic simulation framework.
In designing QPyTorch, we have the following objectives.
First, we aim to support broad and general research into low-precision numeric formats.
Popular machine learning frameworks like Tensorflow~\cite{tensorflow} now support simulating fixed-point computation, in addition to the 16-bit low-precision floating point numbers that are often supported directly on hardware.
However, recent algorithms often propose low-precision number formats that are more general than these options~\cite{dyanmic-fixed-point, Float-8}, including block floating point and non-standard-width floating point numbers. 
To better support this sort of research, we support a diverse range of numeric design choices with our framework, including allowing arbitrary mix-and-match of different precisions, number formats, and rounding options.

Second, we value efficiency on commodity hardware. 
The current state of low-precision training research often requires validations on large-scale, realistic problems (e.g. ImageNet~\cite{imagenet}). 
Efficient simulation enables fast iterations in empirical research.
Unfortunately there is inherent trade-off between speed and generality---in some cases, we eschew support for a low-precision operation that would greatly diminish simulator efficiency.
In this report, we document the scope of the features QPyTorch supports and evaluate our design choices regarding efficiency.

Third, we develop a convenient front-end interface. 
Low-precision simulation usually ``hijacks'' a regular computation graph in deep learning training, and
quantization code can easily be buried within a complex machine learning codebase.
QPyTorch provides proper abstractions to elucidate the high-level designs in low-precision training, which helps understanding and reproducibility.
Additionally, our interface allows users to adjust the settings for individual layer or number categories independently.
QPyTorch also implements a common set of low-precision training techniques from the literature\footnote{For instance, many training algorithms keep a higher-precision model copy for gradient accumulation.}~\cite{binaryconnect, WAGE, Float-8, swalp}, which can serve as a set of building blocks for future exploration.
%QPyTorch implements these techniques.

Finally, we focus on the reliability of QPyTorch.
Naive implementation of low-precision simulation can be error-prone.
For example, high-precision numbers can be accidentally leaked into the training process.
In the past, we have found quantization bugs in low-precision codebases that could potentially invalidate some conclusions in the published literature.
It is our hope that QPyTorch can help prevent this class of error, both because QPyTorch contains standard test suites to validate the implementation, and because potential errors may be exposed more quickly by a growing open-source community working on a single codebase.

\section{Related Works}

\textbf{Low-level low-precision packages.}
Tensorflow-Lite~\cite{tensorflow-quant} supports dynamically quantizing a trained model into 8 bits at inference time and store computation results in floating point.
QNNPACK~\cite{QNNPACK} provides a mobile-optimized 8-bit fixed point implementation.
FBGEMM~\cite{FBGEMM} provides low-level matrix-multiplication and convolution support for half precision floating point and 8-bit fixed point.

\textbf{Low-precision simulation frameworks.}
TensorQuant~\cite{tensorquant} is one of the first software frameworks that supports fixed point computation simulation. 
Tensorflow~\cite{tensorflow} supports simulating fixed point computation of arbitrary bits with nearest rounding.
In comparison, our framework supports a wider range of number formats and rounding options.
QPyTorch also allows the users to configure individual layers and number categories independently.
    
\section{Design}
At its core, QPyTorch operates by (1) representing low-precision numbers as their corresponding floating-point number, and (2) simulating low-precision computation by running single-precision floating-point computation and then removing the extra precision through quantization.
For individual base operations (e.g.\ scalar $+$, $\times$), this approach is equivalent to low-precision computation.
However, for composite operations (e.g.\ matrix multiplication), this corresponds to using a high-precision accumulator rather than accumulating in the same low-precision format: this choice is standard in the literature~\cite{WAGE, swalp}.
For QPyTorch, we are motivated by efficiency: quantizing after each base operation would significantly slow down the simulator (see discussion in \citet{tensorquant}).

\subsection{Features of QPyTorch}
\paragraph{Number Formats.}

\begin{figure}[t!]
    \centering
    \begin{subfigure}[b]{0.65\textwidth}
        \centering
        \includegraphics[width=\textwidth]{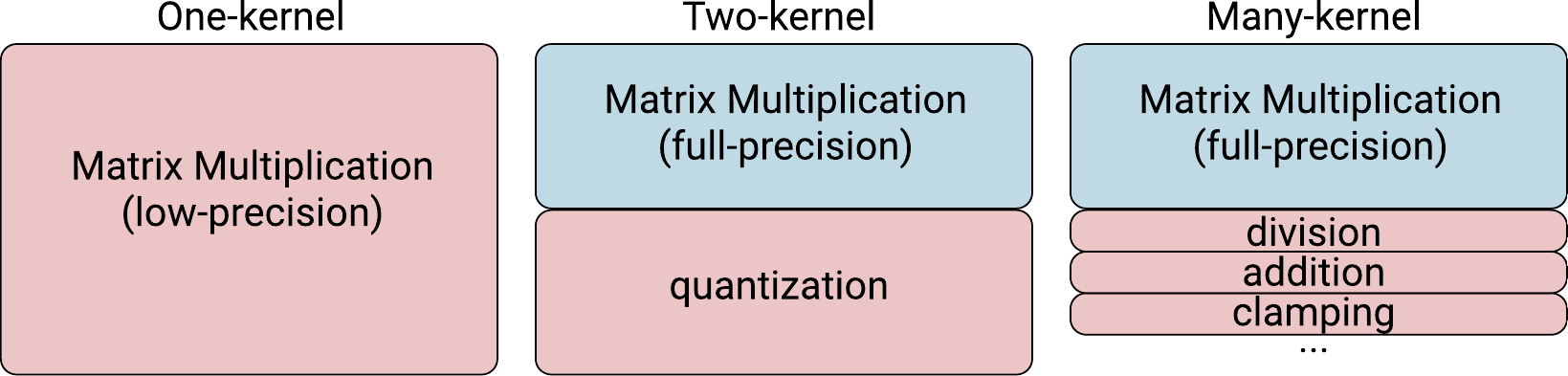}
        \caption{Illustration of QPyTorch's two-kernel approach.}
        \label{fig:approach}
    \end{subfigure}\hfill
    \begin{subfigure}[b]{0.3\textwidth}
        \centering
        \includegraphics[width=\textwidth]{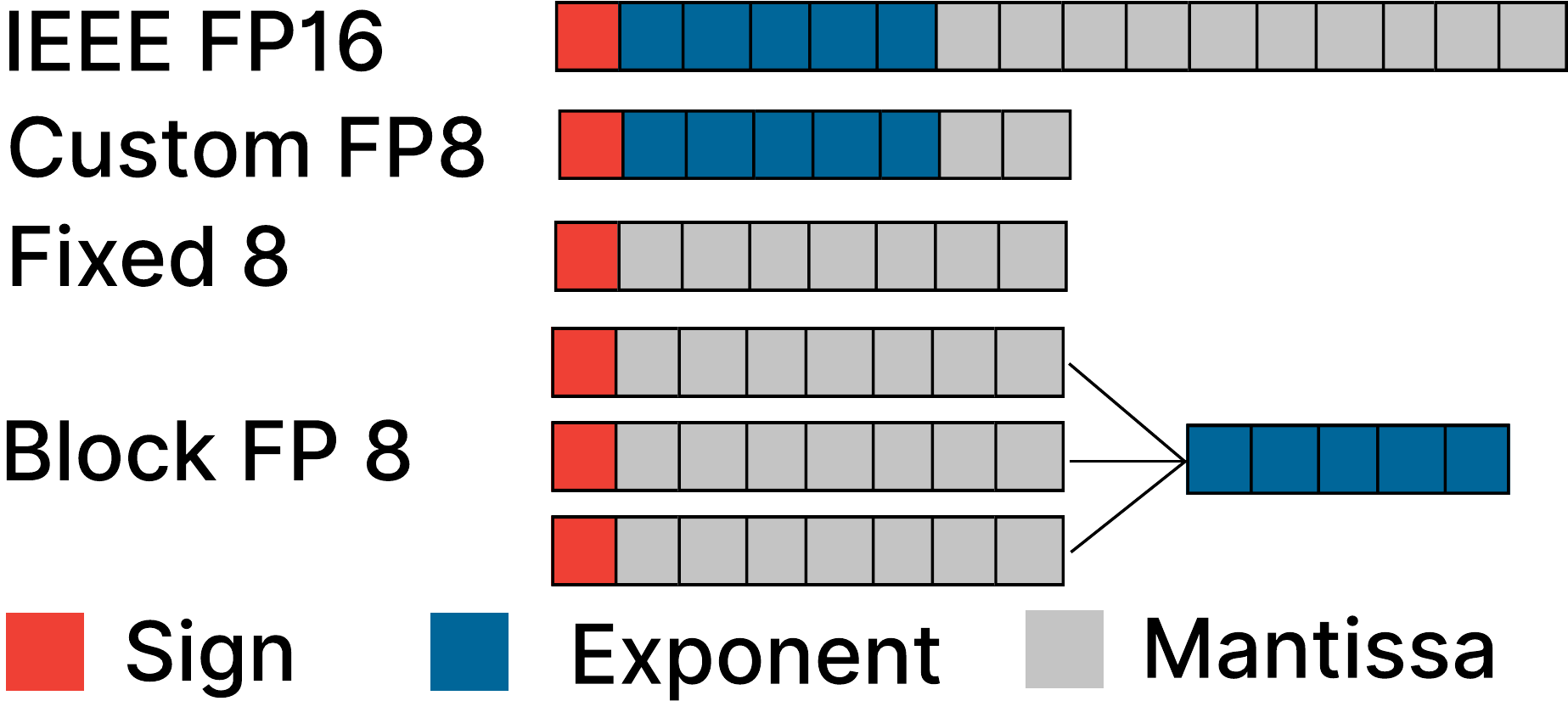}
        \caption{Example supported formats}
        \label{fig:number-format}
    \end{subfigure}
    \caption{}
    
\end{figure}

QPyTorch can simulate a diverse set of low-precision number formats (Fig~\ref{fig:number-format}), including floating point, fixed point, and block floating point numbers. 
For floating point numbers, QPyTorch can simulate any floating point format that uses fewer bits than single precision (i.e. those that use fewer than 8 bits for exponent and fewer than 23 bits for mantissa).
Among these are the Brain Floating Point (bfloat16) and the effective 8-bit and 16-bit format proposed in \citet{Float-8}.
In the interest of efficiency, we do not simulate denormals, NaN, and Inf, as these numbers are expected to occur rarely in training and are often not all supported in low-precision hardware~\cite{intel-bfloat}.

QPyTorch also supports fixed point, which is a popular format that has been used in many algorithms~\cite{gupta2015deep, WAGE}.
QPyTorch can simulate fixed point numbers with arbitrary precision although the precision is again limited by the underlying single precision computation (restricting us to fixed-point numbers with at most 24 bits of precision).
Block floating point numbers reduce the bit usage of floating point numbers by sharing the exponent bits across a block~\cite{swalp}.
QPyTorch can simulate any block floating point format that allocates fewer than 8 bits for exponent and fewer than 23 bits for mantissa.
The assignment of numbers to blocks appears to be an important design decision in low-precision training algorithms that use block floating point~\cite{dorefa-net, swalp}.
QPyTorch supports assigning the block over an arbitrary dimension of a tensor or treating the tensor as an single block.
Finally, QPyTorch provides a unified API to combine these primitive quantizations with additional PyTorch code to define more specialized formats~\cite{WAGE, dorefa-net}.

\paragraph{Rounding mode.}
Stochastic rounding has been shown to be beneficial for neural network training~\cite{gupta2015deep}.
QPyTorch implements both stochastic and nearest-neighbor rounding.
For handling the middle-points with nearest-neighbor rounding, QPyTorch supports round-away-from-zero, round-toward-zero, and round-to-nearest-even.

\paragraph{Front-end interface.}

\begin{figure}[t!]
    \centering
    \begin{subfigure}[t]{0.5\textwidth}
        \centering
        \includegraphics[width=\textwidth]{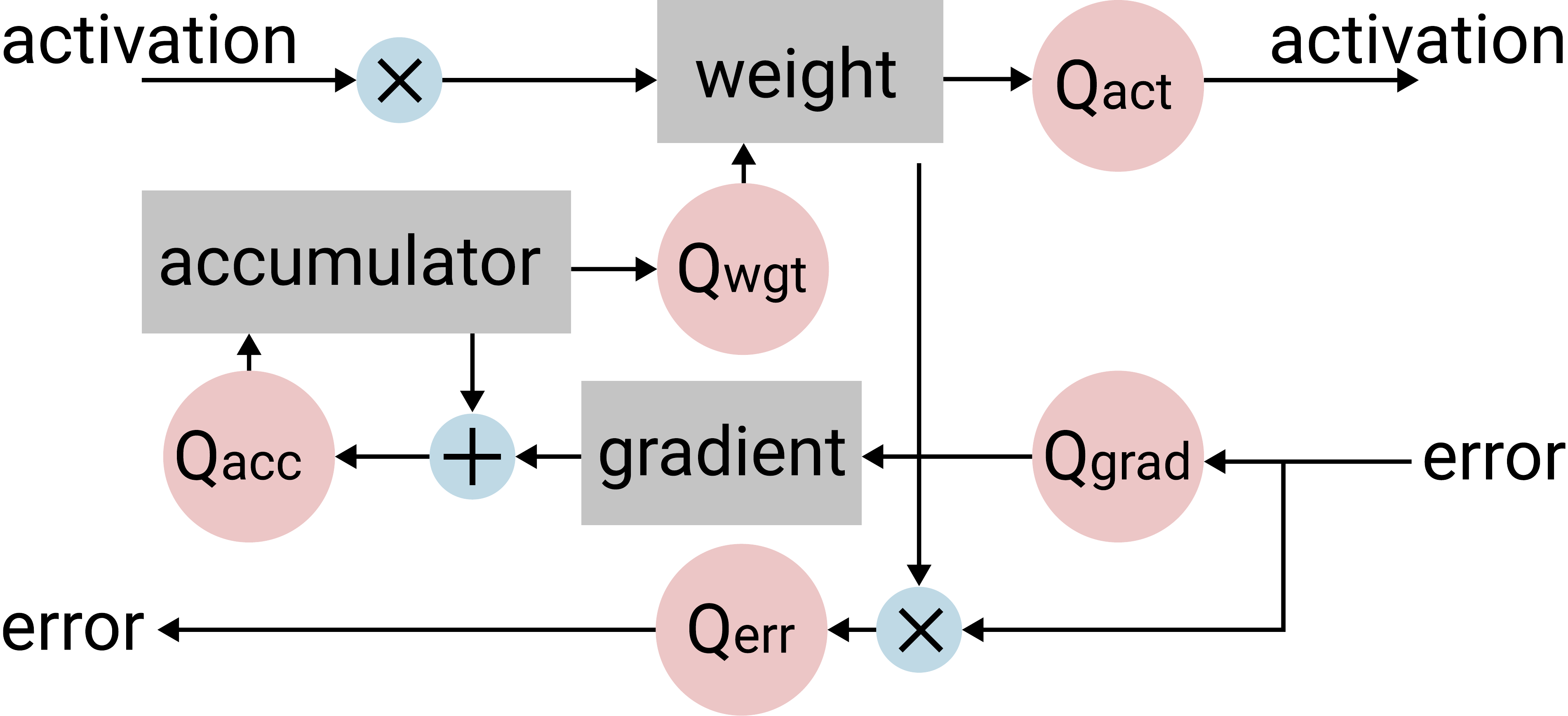}
        \caption{Categories of numbers in low-precision training}
        \label{fig:number-categories}
    \end{subfigure}\hfill
    \begin{subfigure}[t]{0.47\textwidth}
        \centering
        \includegraphics[width=\textwidth]{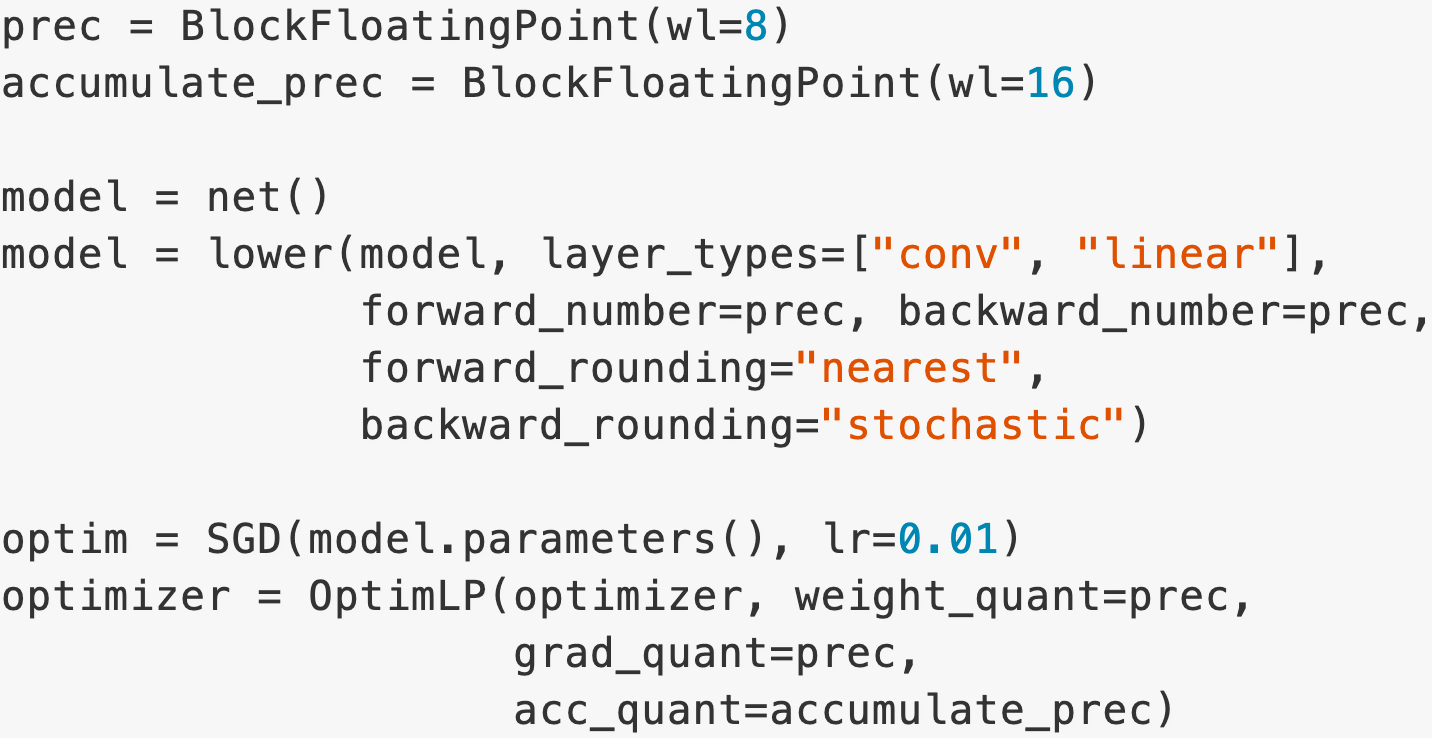}
        \caption{Example QPytorch implementation}
        \label{fig:example-snippet}
    \end{subfigure}
    \caption{}
    
\end{figure}
    
Neural network training involves distinctive groups of numbers~\cite{DMGC}. 
Figure~\ref{fig:number-categories} illustrates the different roles of weight, accumulator, gradient, activation, and error in a single layer. 
Different categories and different layers in a neural network may require distinct precision and quantization~\cite{gupta2015deep}.
QPyTorch provides a convenient interface to handle different number categories independently, allowing the search for more effective combinations of different precision, formats, and rounding mode.
Specifically, QPyTorch injects the quantization of activation and error as a separate PyTorch module into each layer and abstracts the quantization of weight, gradient, and accumulators into a low-precision optimizer.
In addition, we provide a useful tool to automatically inject the quantization modules so that low-precision networks do not require a separate model definition.
Using these convenient designs, it only takes around 10 lines of code to convert an existing full-precision training codebase into a low-precision one (Figure~\ref{fig:example-snippet}).

\subsection{Two-Kernel Approach}
One can imagine several potential designs for QPyTorch. 
The most straightforward is to utilize existing PyTorch operations directly. 
The caveat is that each PyTorch operation would launch a separate CUDA kernel and many separate launches are inefficient.
Moreover, due to the lack of support for bitwise operations in PyTorch, this approach cannot simulate low-precision floating point numbers.
From now on, we refer to this strategy as the \emph{many-kernel approach}.
On the other hand, we could implement a customized kernel for each common operation, such as matrix multiplication, convolution, nonlinearities, etc.
This \emph{one-kernel design} is a popular choice when efficiency is of major concern, e.g., in software packages designed to leverage low-precision hardware~\cite{QNNPACK, FBGEMM}.
Although fast, this approach is cumbersome for general research purposes: 
as any new operations arising in frontier research would require a separate kernel, this adds significant maintenance overhead.
QPyTorch adopts an intermediate approach.
We fuse instructions specific to quantization into a single kernel and then append it to a regular full-precision operation run in a separate kernel.
Figure~\ref{fig:approach} illustrates this \emph{two-kernel approach}.

\section{Validation}
\begin{figure}[!t]
  \begin{minipage}{\textwidth}
  \begin{minipage}[b]{0.32\textwidth}
        \centering
        \resizebox{\textwidth}{!}{  
        \begin{tabular}{l c c}
        \toprule
        Algorithm & PyTorch & QPyTorch  \\
        \midrule
        FP16 SGD & 6.88 &6.89\\
        Block8 SGD & 8.24 & 8.4 \\
        WAGE & 6.96 & 6.95 \\
        SWALP & 6.7 & 6.67 \\
        \bottomrule
        \end{tabular}}
        \captionof{table}{Error rate of four models trained with PyTorch and QPyTorch on CIFAR10.}
        \label{tab:replication}
  \end{minipage}
  \hfill
  \begin{minipage}[b]{0.32\textwidth}
    \centering
    \includegraphics[width=\linewidth]{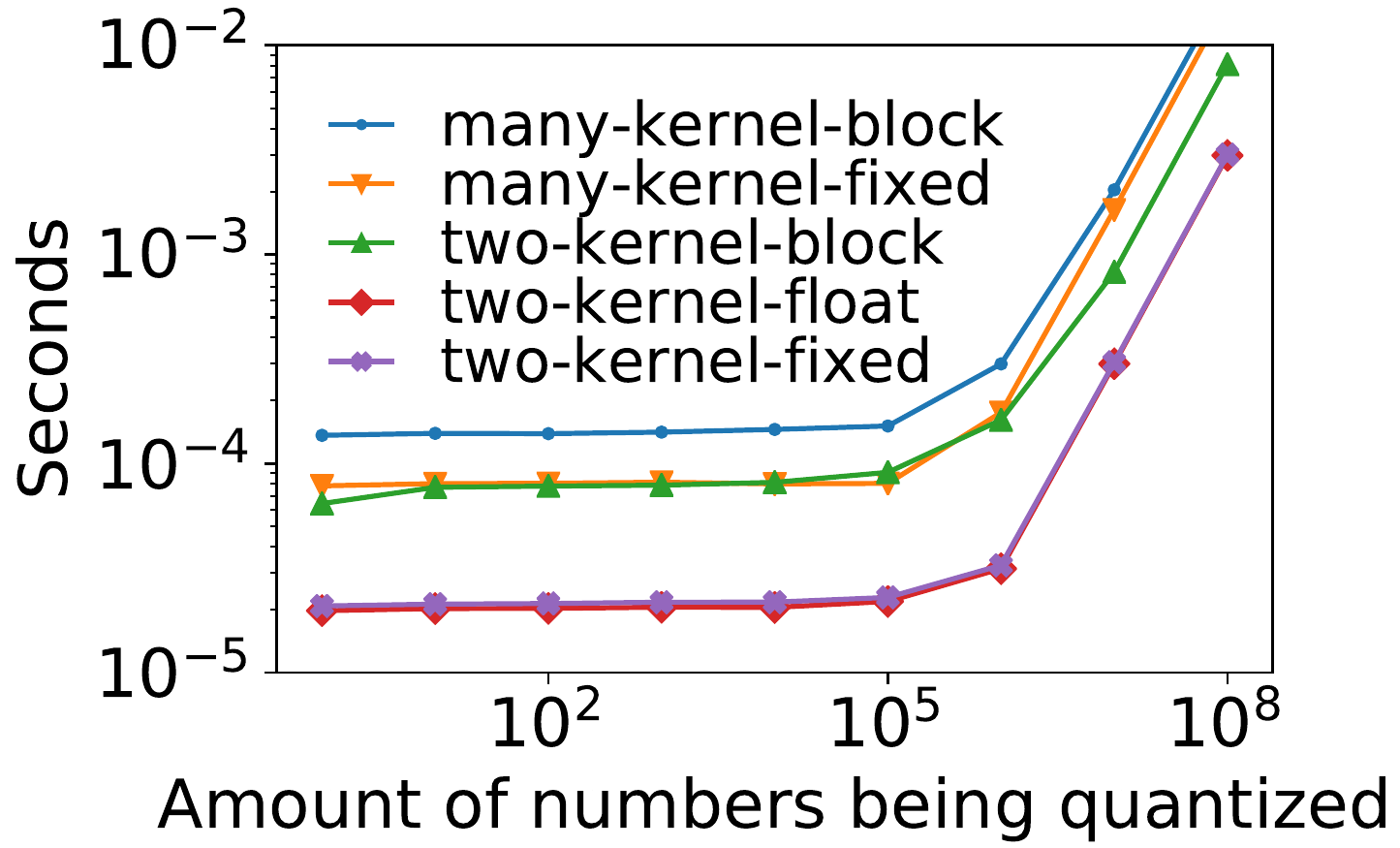}
        \captionof{figure}{Wall Clock Time of Quantization Kernels.
        }
    \label{fig:wall-clock}
    \end{minipage}
  \hfill
  \begin{minipage}[b]{0.32\textwidth}
    \centering
    \includegraphics[width=\linewidth]{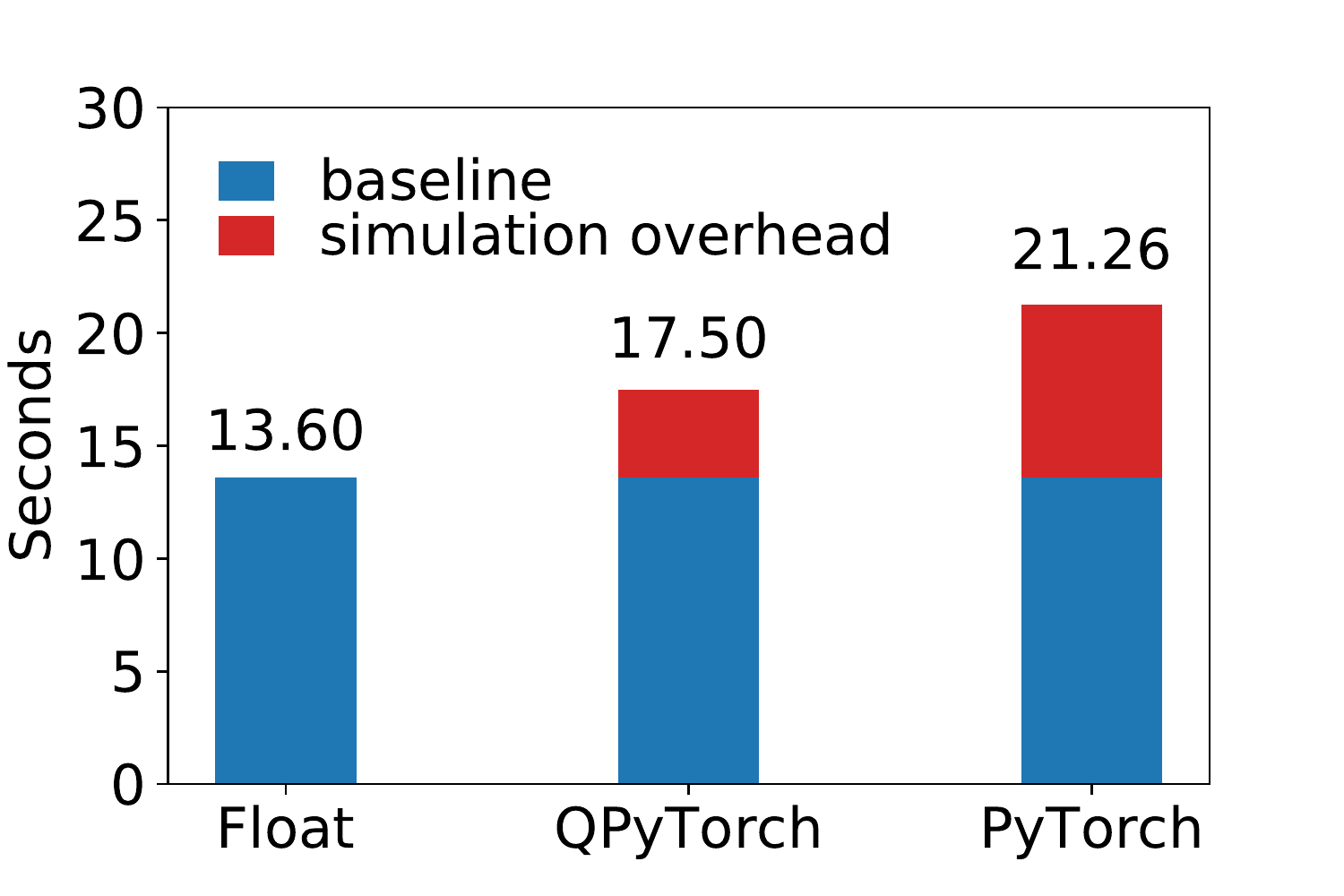}
        \captionof{figure}{Wall Clock Time of training WAGE for one Epoch. 
        }
    \label{fig:overhead}
    \end{minipage}
  \end{minipage}
\end{figure}

To validate the correctness of our framework, we train various low-precision models. 
We use the CIFAR10~\cite{CIFAR10} dataset to benchmark the system performance.
We first study applying stochastic gradient descent to train VGG16 models with two precision settings, half-precision floating point and 8-bit block floating point where each tensor is treated as a block.
We apply standard data augmentation and borrow other training hyperparameters from \citet{swalp}.
Then, we implement two state-of-the-art low-precision training algorithms~\cite{swalp, WAGE} in QPyTorch and compare to the reported performance. 
For SWALP~\cite{swalp}, we train a VGG16 model using the reported 8-bit small-block block floating point format.
For WAGE~\cite{WAGE}, we implement the customized VGG model and the specialized number format, which can be viewed as a variant of fixed point.
Table~\ref{tab:replication} shows that QPyTorch performs as expected. 

We also evaluate QPyTorch's two-kernel approach.
We cannot fairly compare to a one-kernel approach by modifying existing optimized full-precision kernels because source code for these kernels is not publicly available:
therefore, we restrict our evaluation to a comparison between the two-kernel and the many-kernel approaches.
We implemented the many-kernel approach directly through PyTorch operations, and compare it with our CUDA implementation of the two-kernel approach used in QPyTorch.
Since PyTorch does not support bitwise shift operations, a many-kernel implementation of  floating point quantization is not possible, so we restrict our comparison to the fixed point and block floating point formats.
We denote our PyTorch implementation of fixed point quantization \texttt{many-fixed} 
and block floating point quantization \texttt{many-block}. 
Similarly, denote the fused kernels of QPyTorch \texttt{two-kernel-fixed}, \texttt{two-kernel-block}. 
We refer to the floating point quantization in QPyTorch as \texttt{two-kernel-float}. 
In Figure~\ref{fig:wall-clock}, we plot the wall clock time spent quantizing tensors of different sizes. 
% We benchmark the wall clock time on CPU and GPU with both nearest rounding and stochastic rounding.
Figure~\ref{fig:wall-clock} demonstrates the efficiency of our two-kernel approach. 
Notably, \texttt{two-kernel-fixed} with nearest rounding on GPU is about 5x faster than \texttt{many-kernel-fixed}.
In addition, for Block Floating Point which is most expensive to simulate, \texttt{two-kernel-block} is about 2.6x faster than \texttt{many-kernel-block}.
We observe similar results for stochastic rounding and quantizing on CPUs.
Additionally, we compare the run time of training WAGE for one epoch. 
Figure~\ref{fig:overhead} shows that QPyTorch effectively halves the end-to-end simulation overhead (3.9s vs. 7.56s).

\section{Conclusion and Future Work}
In this report we introduce QPyTorch, a low-precision arithmetic simulation framework.
QPyTorch targets low-precision training research, facilitating studies on various number formats, rounding choices.
QPyTorch adopts a two-kernel approach for efficient simulation and offers a convenient interface that is tailored for recent algorithms.
We validate QPyTorch by training low-precision neural networks in four diverse settings and demonstrate its efficency empirically.

\newpage

\bibliographystyle{plainnat}
\bibliography{main}

\end{document}